\documentclass[lettersize,journal]{IEEEtran}
\usepackage{amsmath,amsfonts}
\usepackage{array}
\usepackage[caption=false,font=normalsize,labelfont=sf,textfont=sf]{subfig}
\usepackage{textcomp}
\usepackage{stfloats}
\usepackage{url}
\usepackage{verbatim}
\usepackage{graphicx}
\usepackage{cite}
\usepackage{mathtools}
\usepackage{booktabs}
\usepackage{enumitem}
\usepackage[
colorlinks=true,
linkcolor=blue,
citecolor=blue,
urlcolor=blue,
filecolor=blue
]{hyperref}
\usepackage{cuted} 
\usepackage{caption}
\usepackage{epsfig} 
\usepackage{multirow}
\usepackage{threeparttable}
\usepackage{bm}
\usepackage{multicol}
\usepackage[table,xcdraw]{xcolor}
\usepackage{tikz}
\usepackage{adjustbox}
\usepackage{amssymb}
\usepackage{algorithm}
\usepackage{algpseudocode}
\usepackage{makecell}

\newcommand{\revlzh}[1]{{\color{black}#1}}
\algrenewcommand\algorithmiccomment[1]{\hfill\textcolor{gray}{// #1}}

\begin{document}
	
	\title{SGTP: Sampling-based Game-Theoretic Planning\\for Real-Time Multi-Vehicle Autonomous Racing}
	
	\author{Zhouheng Li$^{1,2}$, Fangguo Zhao$^{1}$, Mattia Piccinini$^{3}$, Baha Zarrouki$^{3}$, Yuan Gao$^{3}$,  Zitong Shan$^{2}$,\\ Johannes Betz$^{3}$, Chen Lv$^{2}$, Lei Xie$^{1,\dagger}$
		\thanks{$^{\dagger}$Corresponding author: leix@iipc.zju.edu.cn.}
		\thanks{$^{1}$Zhejiang University, Hangzhou, China. $^{2}$Nanyang Technological University, Singapore. $^{3}$Technical University of Munich, Garching, Germany.  }
		\vspace{-1.1cm}
	}
	
	\maketitle
	\renewcommand{\thefootnote}{\alph{footnote}}

	\begin{abstract}
		Autonomous multi-vehicle racing requires real-time planning of diverse competitive behaviors in intense interactions. Existing planners often struggle to balance strategic diversity and computational efficiency. 
		To address this challenge, we propose Sampling-based Game-Theoretic Planning (SGTP), a real-time framework that combines game-theoretic reasoning with GPU-accelerated sampling of control sequences and dynamics rollouts. Sampled trajectories are ranked using a game-aware cost to capture competitive interactions and generate diverse racing behaviors. Our planner then performs feasibility selection by explicitly enforcing track-boundary and dynamic collision-avoidance constraints, ensuring safe and reliable transitions between racing strategies.
		Extensive simulations on challenging tracks show that SGTP achieves a 95.24\% win rate and a 99.35\% task-completion ratio in highly interactive races, with a mean computational time of 0.095 s over multiple iterative solving steps. We also demonstrate the successful application of SGTP in large-scale scenarios with up to 10 agents. We release our code and provide an open-source benchmark of multi-agent autonomous racing algorithms to facilitate future research. Project page: \url{https://sgtp-racing.github.io/}.
	\end{abstract}
	
	\vspace{-10pt}

	\section{Introduction}
	
	In Formula 1 racing, drivers continuously manage gaps to create overtaking opportunities and defend their positions. Reproducing such sustained wheel-to-wheel interaction remains challenging in autonomous racing, as it requires planners to generate multiple competitive behaviors and transition reliably between them as interactions evolve.

	Despite extensive research on rule-based, optimization-based, sampling-based, and learning-based trajectory planning methods, generating competitive behaviors in multi-vehicle racing remains challenging~\cite{Betz2022_survey}. 
	Finite-State-Machine (FSM)-based  planners~\cite{baumann2025forzaeth,toschi2025modular} have shown effective multi-behavior planning, but they rely on expert-designed behavior modes and heavy parameter tuning.  
	Model Predictive Contouring Control (MPCC)~\cite{liniger2015optimization,10920215,li2025data}  and Model Predictive Control (MPC)~\cite{Piccinini2025} explicitly model racing progress and reference velocity profiles. However, close-proximity interactions make the resulting optimization problems sensitive to initialization, leading to variable computational time. 
	Sampling-based planners \cite{Langmann2026} can provide an effective trade-off between domain exploration and computational efficiency, but existing methods are limited to 1-2 opponent vehicles. 
	Learning-based methods, including end-to-end policies~\cite{qiao2025end2race} and generative models~\cite{NEURIPS2025_36d1e8aa,zheng2025diffusion}, can represent rich interaction behaviors, but their limited interpretability and cross-scenario generalization remain concerns~\cite{mao2025rapid}. These limitations motivate new planning methods that leverage interaction-aware reasoning to explore diverse strategies while ensuring reliable transitions between them.
	
	\begin{figure}[!t]
		\centering
		{\includegraphics[width=3.5in]{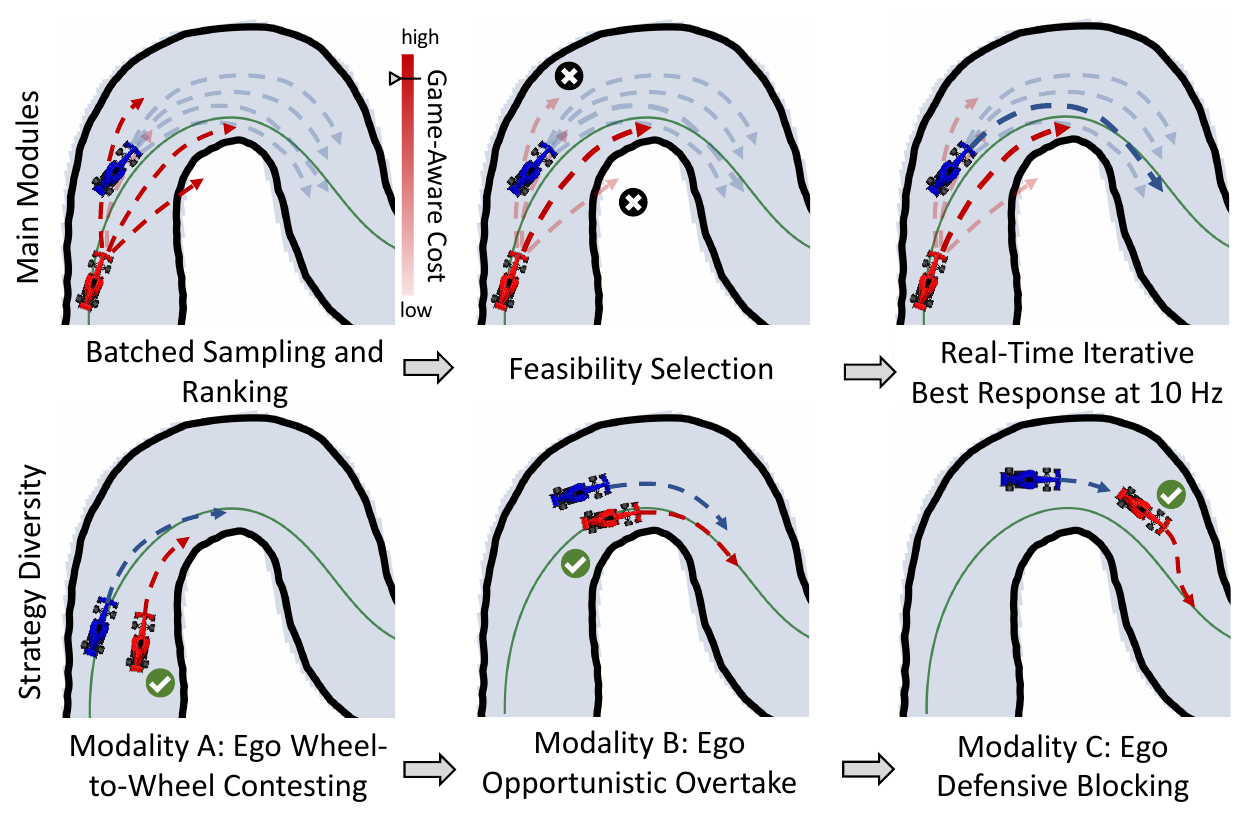}}
		\caption{\textbf{Proposed SGTP framework for multi-behavior autonomous racing.} SGTP combines GPU-accelerated sampling-based planning with game-theoretic best-response reasoning, using a new game-aware cost to favor competitive interactions and selecting the lowest-cost feasible trajectory.}
		\label{fig:teaser}
		\vspace{-5pt}
	\end{figure}

	Game theory provides a principled framework for interaction-aware reasoning and multi-strategy planning~\cite{papuc2026strategizing,kalaria2024alpha,pmlr-v283-kim25a}. Accordingly, multi-vehicle racing can be formulated as a Generalized Nash Equilibrium Problem~\cite{distelzweig2025drivinggamecombiningplanning,prignoli2025regulation,10160799} and is commonly solved using Iterative Best Response (IBR), which  optimizes each vehicle’s strategy while holding those of the other vehicles fixed~\cite{spica2020real,wang2021game,prignoli2025regulation}. 
	However, consistently obtaining feasible and high-quality solutions to each best-response subproblem within the available online computation time remains a major bottleneck~\cite{10160799,prignoli2025regulation}. Although MPC is widely adopted for its ability to explicitly handle system and safety constraints~\cite{wang2021game,prignoli2025regulation}, the resulting optimization problems are highly nonconvex~\cite{wang2021game}; consequently, both solution quality and computational time are highly sensitive to initialization~\cite{LI2024104816}, particularly in close-range racing scenarios~\cite{wang2021game}. As a result, MPC-based best-response solvers exhibit initialization-dependent or unpredictable convergence, and impose substantial online computational burdens~\cite{prignoli2025regulation,Langmann2026}.
	Model Predictive Path Integral (MPPI)-based planners~\cite{honda2025model,trevisan2024biased,honda2024stein} have emerged as GPU-accelerated sampling-based alternatives for real-time trajectory planning. However, they typically handle collision avoidance through soft cost penalties rather than explicit feasibility checks, which can lead to infeasible maneuvers in close-range wheel-to-wheel racing. Despite the potential of integrating MPPI with IBR and tailored game-theoretic cost functions to alleviate existing limitations, this combination remains unexplored.

	We propose a \textbf{S}ampling-based \textbf{G}ame-\textbf{T}heoretic \textbf{P}lanning (\textbf{SGTP}) framework for competitive multi-vehicle racing that generates diverse racing strategies in real time. SGTP integrates IBR by reformulating each best-response sub-problem as GPU-parallel control sampling, followed by dynamics rollout and trajectory ranking. A game-aware cost encourages competitive interactions, while feasibility selection explicitly enforces track-boundary and collision-avoidance constraints, enabling reliable transitions between strategies (Fig.~\ref{fig:teaser}). Existing evaluation frameworks often rely on simplified maps and lack interaction-rich benchmarks, limiting the assessment of planners in highly competitive racing scenarios~\cite{wang2021game,10160799}. To address this gap, we release our SGTP planner, its benchmarks and the evaluation platform as open source. The main contributions of this letter are summarized as follows:

	\begin{enumerate}
		\item \textbf{SGTP: A Real-Time, Sampling-Based, IBR Trajectory Planning Framework.}
		SGTP performs GPU-parallel batched rollout generation and cost-based selection to approximate unilateral best responses, followed by an additional ego response after opponent prediction. Our design enables multiple competitive behaviors in intense interactions, while achieving real-time planning with a mean computation time of \textbf{0.095s}.
		\item \textbf{Game-Aware Trajectory Cost with  Feasibility Selection.}
		A new game-aware cost evaluates rollout trajectories based on relative progress, contest status, blocking alignment, and safety margins, thereby promoting diverse competitive behaviors. Combined with feasibility selection over the ranked rollouts, SGTP achieves a  \textbf{95.24\%} win rate and a \textbf{99.35\%} completion ratio in long-duration competitive racing.
		\item \textbf{Open-Source Benchmark of Multi-Agent Autonomous Racing.}
		We release an open benchmark with SGTP, multiple literature baselines, and several track layouts, to evaluate trajectory planning methods in multi-vehicle autonomous racing.  
	\end{enumerate}
	
	\vspace{-10pt}

	\section{SGTP Algorithm}
	\label{sec:problem_formulation}
	\paragraph{Preliminaries and Definitions}
	Let $\mathcal{I}=\{1,\ldots,n_v\}$ denote the set of racing vehicles, where $n_v$ is the number of vehicles. For each vehicle $i\in\mathcal{I}$, $\mathcal{I}_{-i}=\mathcal{I}\setminus\{i\}$ denotes the set of opponent vehicles.
	Let $t$ denote the current discrete planning step. The joint multi-vehicle state at time $t$ is defined as $X_t=\{\mathbf{x}_{i,t}\}_{i\in\mathcal{I}}$, where $\mathbf{x}_{i,t}=[x_{i,t},y_{i,t},\psi_{i,t},v_{i,t}]^\top$ and $\mathbf{u}_{i,t}=[a_{i,t},\delta_{i,t}]^\top$ denote the state and control input, respectively. 
	Here, $x_{i,t}$ and $y_{i,t}$ are the vehicle position, $\psi_{i,t}$ is the heading angle, $v_{i,t}$ is the speed, $a_{i,t}$ is the longitudinal acceleration, and $\delta_{i,t}$ is the steering angle. The vehicle dynamics are written as $\mathbf{x}_{i,t+1}
	=
	\boldsymbol{f}(\mathbf{x}_{i,t},\mathbf{u}_{i,t}),$ where $\boldsymbol{f}(\cdot)$ follows the kinematic bicycle model in~\cite{7995816}. 
	The sampled control sequence  $\boldsymbol U_{i,t}
	$ is $
	[
	\mathbf u_{i,t},
	\ldots,
	\mathbf u_{i,t+H-1}
	]$, and the corresponding trajectory $\boldsymbol{\xi}_{i,t}$ is denoted by 
	$
	\left[
	\mathbf{x}_{i,t},
	\ldots,
	\mathbf{x}_{i,t+H}
	\right]$, where  $H$ is the prediction horizon.
	For the  opponent set $\mathcal{I}_{-i}$, the predicted trajectories are  $\widehat{\boldsymbol{\Xi}}_{-i,t}=\{\hat{\boldsymbol{\xi}}_{j,t}\}_{j\in\mathcal{I}_{-i}}$, where $\hat{\boldsymbol{\xi}}_{j,t}$ denotes the predicted trajectory of vehicle $j$.
	For a given racetrack, we compute a minimum-curvature raceline
	$
	\mathcal{G}
	=
	\left\{
	(\mathbf{x}_j^r,w_j^L,w_j^R)
	\right\}_{j=1}^{N_{\mathrm{trk}}},
	$
	where $N_{\mathrm{trk}}$ is the number of waypoints, $\mathbf{x}_j^r$ is the raceline state trajectory, $w_j^L$ and $w_j^R$ are the left and right track widths. The raceline generation follows~\cite{heilmeier2020minimum}. By interpolating these arc-length-parameterized waypoints, $\mathcal{G}$ defines continuous reference maps $\mathbf{x}^r(s)$, $w^L(s)$, and $w^R(s)$, which are used for reference tracking and track-boundary evaluation. The Frenet progress $s_{i,t}$ along the raceline and lateral offset $d_{i,t}$ of vehicle $i$ are then computed as:
	$
	s_{i,t}=s(\mathbf{x}_{i,t};\mathcal{G}),
	\,
	d_{i,t}=d(\mathbf{x}_{i,t};\mathcal{G}).
	$

	\newcommand{\algcomment}[1]{\hspace{0.3em}{\color{gray}$\triangleright$~#1}}
	\newcommand{\algcommentpurple}[1]{\hspace{0.3em}{\color{purple}$\triangleright$~#1}}
	\begin{algorithm}[!t]
		\caption{Sampling-Based Game-Theoretic Planning}
		\label{alg:SGTP_flow}
		\begin{algorithmic}[1]
			\State \textbf{Inputs:} Current states $\{\mathbf{x}_{i,t}\}_{i\in\mathcal{I}}$, raceline $\mathcal{G}$, horizon $H$, samples $K$, IBR iterations $L_{\mathrm{IBR}}$, ego vehicle index $e$
			\State \textbf{Outputs:} Planned trajectories for all agents $\widehat{\boldsymbol{\Xi}}=\{\boldsymbol{\xi}_i^\star\}_{i\in\mathcal{I}}$
			\State Initialize $\{\boldsymbol{\xi}_{i}^{0}\}_{i\in\mathcal{I}}$ without game-aware costs. \algcomment{Distance-based obstacle avoidance} \label{alo:init_trajs}
			\For{$\ell=1,\ldots,L_{\mathrm{IBR}}$} \label{alo:ibr_loop}
			\For{each vehicle $i\in\mathcal{I}$}  \label{alo:vehicle_loop}
			\State Set $\widehat{\boldsymbol{\Xi}}_{-i}^{\ell}\gets\{\hat{\boldsymbol{\xi}}_{j}^{\ell-1}\}_{j\in\mathcal{I}_{-i}}$ \algcomment{Opponent predictions}\label{alo:set_opp_preds}
			\State Generate $\mathcal C_{i}^{\ell}=\{(\boldsymbol\xi_{i,t}^{(k,\ell)}, \boldsymbol U_{i,t}^{(k,\ell)})\}_{k=1}^{K}$ \label{alo:sampling}
			\State Compute cost $c_i^{(k,\ell)}$ via~\eqref{eq:allc} $\forall$ candidate $k$ in $\mathcal C_{i}^{\ell}$ 
			\label{alo:game_costs}
			\State Compute feasibility indicators $h_i^{(k,\ell)}$ via~\eqref{eq:selection} 
			\label{alo:feasible_ind}
			\State Apply selection $k_i^{\star,\ell}\gets\arg\min_{k:h_i^{(k,\ell)}=1} c_i^{(k,\ell)}$ \label{alo:selection}
			\State Update $\boldsymbol{\xi}_{i}^{\ell}\gets\boldsymbol{\xi}_{i}^{(k^{\star,\ell})}$ and $\overline{\boldsymbol U}_{i}^{\ell+1}
			\leftarrow
			\boldsymbol U_{i}^{(k^{\star,\ell})}$ \label{alo:update}
			\EndFor
			\EndFor
			\State $k_e^* = \arg\min_{k : h_e^{(k)}=1} c_e^{(k)}$ \algcomment{Extra ego response}
			\label{alo:ego_extra_update}
		\end{algorithmic}
	\end{algorithm}
	
	\vspace{0.1cm}
	\paragraph{Algorithm Formulation}
	Algorithm~\ref{alg:SGTP_flow} summarizes our SGTP framework. The trajectories of all vehicles are initialized using simple distance-based collision avoidance (line~\ref{alo:init_trajs}). The algorithm loops over the IBR iterations and vehicles (lines \ref{alo:ibr_loop}-\ref{alo:vehicle_loop}). At each IBR iteration, every vehicle $i\in \mathcal{I}$ computes an approximate best response, while the predicted trajectories of its opponents are held fixed at their preceding-iteration solutions (line~\ref{alo:set_opp_preds}). To compute the best response, we sample $K$ candidate control sequences and propagate them in parallel through the vehicle dynamics (line~\ref{alo:sampling}). 
	Let 
	$
	\overline{\boldsymbol U}_{i,t}^{\ell}
	$=$
	[
	(\overline{\mathbf u}_{i,t}^{\ell})^\top,
	\ldots,
	(\overline{\mathbf u}_{i,t+H-1}^{\ell})^\top
	]^\top
	$  
	denote the nominal control sequence of vehicle $i$ at IBR iteration $\ell$, used as a reference for candidate sampling. $\overline{\boldsymbol U}_{i,t}^{\ell}$ is initialized to zero upon reset and then warm-started with the control sequence $\overline{\boldsymbol U}_{i,t}^{\ell-1}$ from the previous iteration. 
	For each of the $K$ candidate trajectories and each prediction step $\tau$, an independent Gaussian perturbation is sampled as $\boldsymbol\epsilon_{i,t+\tau}^{(k,\ell)}
	\sim
	\mathcal N
	\left(
	\mathbf 0,
	\boldsymbol\Sigma_u
	\right),
	\boldsymbol\Sigma_u
	=
	\operatorname{diag}
	\left(
	\sigma_a^2,\sigma_\delta^2
	\right),$ where $\sigma_a$ and $\sigma_\delta$ are the standard deviations of the acceleration and steering perturbations. The perturbed control input is generated and projected onto an admissible control set $\mathcal U$:
	$$
	\mathbf u_{i,t+\tau}^{(k,\ell)} = \Pi_{\mathcal U} \left( \widetilde{\mathbf u}_{i,t+\tau}^{(k,\ell)} \right),\; \text{where} \;\;
	\widetilde{\mathbf u}_{i,t+\tau}^{(k,\ell)}
	=
	\overline{\mathbf u}_{i,t+\tau}^{\ell}
	+
	\boldsymbol\epsilon_{i,t+\tau}^{(k,\ell)}.
	$$
	Here $\Pi_{\mathcal U}(\cdot)$ applies clipping to  acceleration and steering limits. Starting from $\mathbf{x}_{i,t}^{(k,\ell)}=\mathbf{x}_{i,t}$, we propagate all sampled control sequences in parallel through the dynamics $\boldsymbol{f}(\cdot)$, generating the finite rollout set
	$\mathcal C_{i,t}^{\ell}=\{(\boldsymbol\xi_{i,t}^{(k,\ell)}, \boldsymbol U_{i,t}^{(k,\ell)})\}_{k=1}^{K}$ (line~\ref{alo:sampling}).
	
	Given the fixed opponent predictions $\widehat{\boldsymbol{\Xi}}_{-i}^{\ell}$, the sampled rollouts are ranked by our game-aware cost (described in Sec.~\ref{sec:game_cost}) and checked for feasibility to enforce track-boundary and collision-avoidance constraints (lines~\ref{alo:game_costs}-\ref{alo:feasible_ind}). The lowest-cost feasible rollout is selected in line~\ref{alo:selection} as:
	\begin{equation}
		k_i^{\star,\ell}
		=
		\arg\min_{k:h_i^{(k,\ell)}=1}
		c_i
		\left(
		\boldsymbol\xi_{i,t}^{(k,\ell)},
		\widehat{\boldsymbol\Xi}_{-i,t}^{\ell},
		\mathcal G
		\right),
	\end{equation}
	Here, $k_i^{\star,\ell}$ denotes the selected candidate index, and $h_i^{(k,\ell)}=1$ indicates that candidate $k$ satisfies both track-boundary and dynamic vehicle collision-avoidance constraints.
	
	The selected trajectory and control sequence warm start the next IBR iteration (line~\ref{alo:update}).  After the prescribed number of iterations $L_{\mathrm{IBR}}$, the ego vehicle performs one additional best-response update with the final opponents' predictions (line~\ref{alo:ego_extra_update}).

	\vspace{-5pt}
	
	\section{Sampling-Based Game-Theoretic Planning
	}
	\label{sec:method}
	
\begin{figure*}[!t]
		\centering
		\includegraphics[width=7.0in]{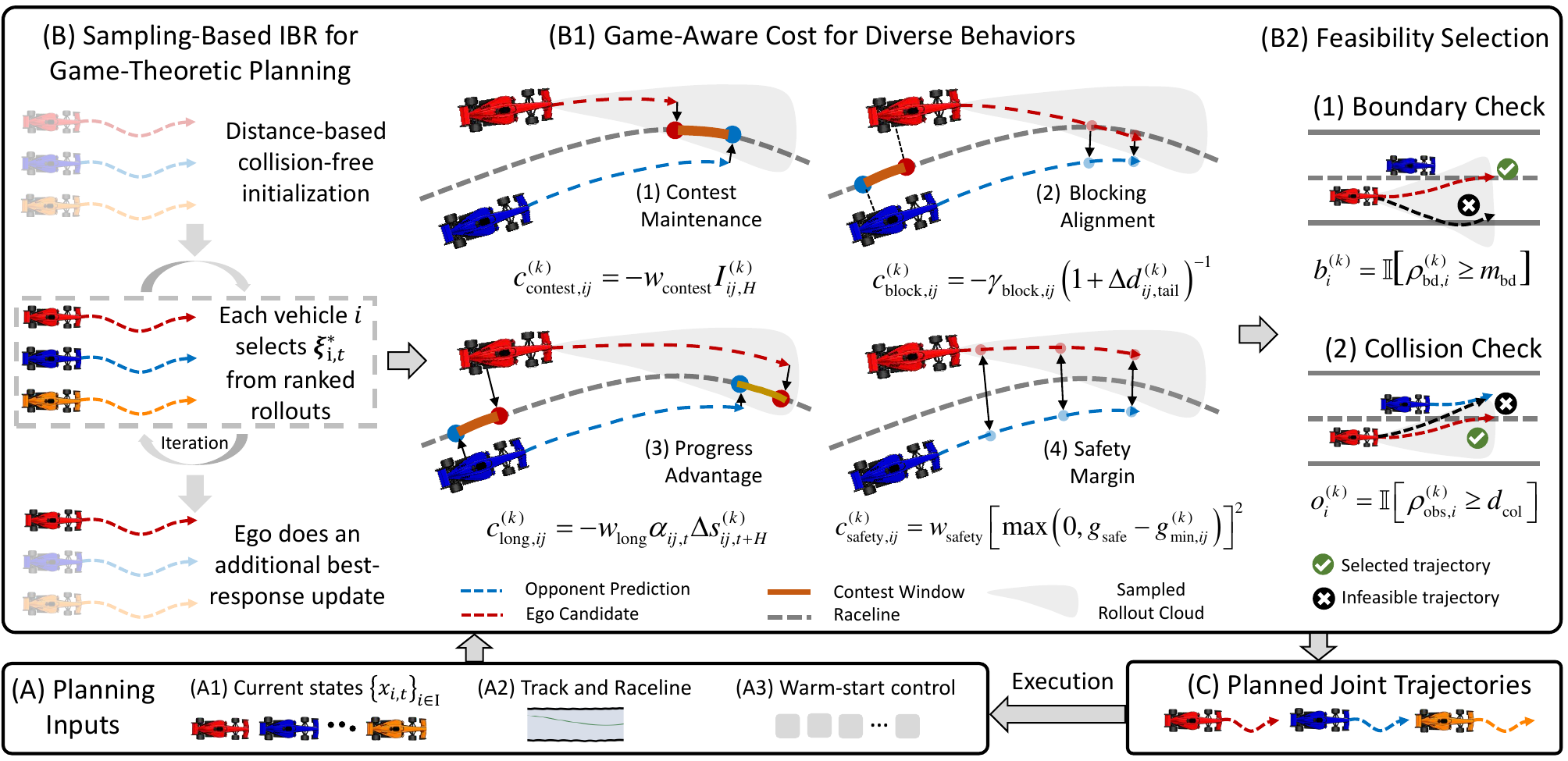}
		\caption{ \textbf{Proposed Sampling-Based Game-Theoretic Planning (SGTP) framework.} SGTP first generates sampled rollouts and ranks them using a game-aware cost evaluated against opponent predictions. During IBR, each vehicle computes a best response, after which feasibility selection enforces the track-boundary and inter-vehicle collision-avoidance constraints. The ego vehicle then performs an additional best-response update.
		}
		\label{fig:pipeline}
		\vspace{-15pt}
	\end{figure*}

	As shown in Fig.~\ref{fig:pipeline}, SGTP performs best-response planning by ranking rollouts (generated in parallel on the GPU) using a game-aware cost, and selecting the lowest-cost feasible rollout.
	\vspace{-20pt}

	\subsection{Game-Aware Cost for Diverse Behaviors}\label{sec:game_cost}

	We compute our game-aware cost by evaluating a pairwise interaction term between each candidate trajectory and the predicted opponent trajectories. For the candidate trajectory $\boldsymbol{\xi}_{i,t}^{(k,\ell)}$ of vehicle $i$ and the predicted trajectory of opponent $j$, the pairwise game-aware cost is defined as follows:
	\begin{equation}\label{eq:game_cost}
		c_{\mathrm{game},ij}^{(k)}
		=
		c_{\mathrm{contest},ij}^{(k)}
		+
		c_{\mathrm{long},ij}^{(k)}
		+
		c_{\mathrm{block},ij}^{(k)}
		+
		c_{\mathrm{safety},ij}^{(k)} .
	\end{equation}
	The four cost terms represent contest-state maintenance $c_{\mathrm{contest}}$, longitudinal advantage $c_{\mathrm{long}}$, blocking alignment $c_{\mathrm{block}}$, and safety separation $c_{\mathrm{safety}}$, respectively. 
	
	The key design insight is to  progressively shape the interaction through complementary cost terms: the contest term preserves rollouts that remain within a meaningful racing window, the longitudinal term promotes forward advantage, the blocking term encourages defensive positioning, and the safety term penalizes insufficient separation.
	The total game-aware cost for candidate $k$ is obtained by summing the pairwise costs over all opponents:
	$
	c_{\mathrm{game},i}^{(k)}
	=
	\sum_{j\in\mathcal{I}_{-i}}
	c_{\mathrm{game},ij}^{(k)},
	$
	allowing each opponent prediction to independently contribute to the ranking of candidate trajectories.

	\subsubsection{Contest-State Maintenance Cost}
	
	The contest-state term rewards candidate trajectories whose terminal progress gaps relative to the predicted opponent trajectories remain within the longitudinal interaction range. Let $s_{i,t+H}^{(k)}$ denote the terminal Frenet progress of ego candidate $k$, and let $\hat{s}_{j,t+H}$ denote the corresponding predicted progress of opponent $j$. The pairwise terminal progress gaps are defined as:
	$
	\Delta s_{ij,t+H}^{(k)}
	=
	s_{i,t+H}^{(k)}
	-
	\hat{s}_{j,t+H}.
	$
	Based on these gaps, the contest indicator for candidate $k$ and opponent $j$ is:
	$$
	I_{ij,H}^{(k)}
	=
	\mathbb{I}
	\left[
	|\Delta s_{ij,t+H}^{(k)}|<s_{\mathrm{contest}}
	\right],
	$$
	where $s_{\mathrm{contest}}$ denotes the longitudinal contest range. The corresponding pairwise contest-state cost is:
	\begin{equation}\label{eq:c_contest}
		c_{\mathrm{contest},ij}^{(k)}
		=
		-
		w_{\mathrm{contest}}
		I_{ij,H}^{(k)}.
	\end{equation}
	where $w_{\mathrm{contest}}$ is a tunable weight. The core insight is to encourage sampled  rollouts to interact closely with opponents’ predicted trajectories within a meaningful competitive window.

	\subsubsection{Longitudinal Advantage Cost}
	
	The longitudinal advantage term encourages candidate trajectories that achieve a terminal progress advantage over opponent $j$. It is defined as:
	\begin{equation}\label{eq:c_long}
		c_{\mathrm{long},ij}^{(k)}
		=
		-
		w_{\mathrm{long}}
		\alpha_{ij,t}
		\Delta s_{ij,t+H}^{(k)},
	\end{equation}
	where $w_{\mathrm{long}}$ is a scalar weight, and 
	$\Delta s_{ij,t+H}^{(k)}$ denotes the terminal progress gap. 
	A larger positive terminal gap reduces the cost, thereby favoring candidates that move ahead of the opponent.
	The adaptive weight $\alpha_{ij,t}$ adjusts this progress reward according to the current interaction relevance:
	$$
	\alpha_{ij,t}
	=
	\left(
	1+
	|\Delta s_{ij,t}|
	(s_{\mathrm{contest}}+\epsilon)^{-1}
	\right)^{-1}.
	$$
	Here $\epsilon$ is a small positive constant. 
	A larger $\alpha_{ij,t}$ assigns greater importance to terminal progress gains when the two vehicles are close.  As their progress gap increases, $\alpha_{ij,t}$ decreases the reward due to weaker interaction relevance.

	\subsubsection{Blocking Alignment Cost}
	
	The blocking alignment term encourages candidate trajectories whose lateral positions align with the predicted position of opponent $j$ near the end of the prediction horizon.
	Let $\mathcal{T}_{\mathrm{tail}}$ denote the set of horizon indices within the final $\mu_{\textrm{tail}}$ portion of the prediction horizon.  The mean lateral offsets of candidate $k$ and opponent $j$ over this final portion of the horizon are defined as follows:
	$$
	\bar{d}_{i}^{(k)}
	=
	\frac{1}{|\mathcal{T}_{\mathrm{tail}}|}
	\sum_{\tau\in\mathcal{T}_{\mathrm{tail}}}
	d_{i,t+\tau}^{(k)},
	\;
	\bar{d}_{j}
	=
	\frac{1}{|\mathcal{T}_{\mathrm{tail}}|}
	\sum_{\tau\in\mathcal{T}_{\mathrm{tail}}}
	\hat{d}_{j,t+\tau},
	$$
	where $\hat{d}_{j,t+\tau}$ denotes the predicted lateral offset of opponent $j$. 
	The corresponding tail lateral gap is:
	$
	\Delta d_{ij,\mathrm{tail}}^{(k)}
	=
	\left|
	\bar{d}_{i}^{(k)}
	-
	\bar{d}_{j}
	\right|.
	$
	Using this gap, the blocking alignment cost is formulated as:
	\begin{equation}\label{eq:c_block}
		c_{\mathrm{block},ij}^{(k)}
		=
		-
		\gamma_{\mathrm{block},ij}
		\left(
		1+
		\Delta d_{ij,\mathrm{tail}}^{(k)}
		\right)^{-1}.
	\end{equation}
	Here, $\small
	\gamma_{\mathrm{block},ij}
	=
	w_{\mathrm{block}}\mathbb{I}\!\left[
	\Delta s_{ij,t}>s_{\mathrm{role}}
	\right]
	\mathbb{I}\!\left[
	\Delta s_{ij,t}<s_{\mathrm{contest}}
	\right]$ activates the blocking cost when vehicle $i$  has a longitudinal advantage of at least $s_{\mathrm{role}}$  while remaining within the contest range $s_{\mathrm{contest}}$, where $\Delta s_{ij,t}
	=
	s_{i,t}-\hat{s}_{j,t}$. The scalar $w_{\mathrm{block}}$ is the  weight.
	Such lateral alignment reduces the space available to the trailing vehicle and promotes defensive blocking.

	\subsubsection{Safety Cost}
	
	The safety term penalizes candidate rollouts whose predicted separation from an opponent falls below the desired threshold $g_{\mathrm{safe}}$. 
	For candidate $k$ and opponent $j$, the minimum time-aligned distance is:
	\begin{equation}\label{eq:Dmin}
		g_{\min,ij}^{(k)}
		=
		\min_{\tau=1,\ldots,H}
		\left\|
		\mathbf{p}_{i,t+\tau}^{(k)}
		-
		\hat{\mathbf{p}}_{j,t+\tau}
		\right\|_2 ,
	\end{equation}
	where $\mathbf{p}_{i,t+\tau}^{(k)}$ and $\hat{\mathbf{p}}_{j,t+\tau}$ denote the position associated with candidate $k$ and the predicted position of opponent $j$ at time $t+\tau$, respectively. The corresponding pairwise safety cost is:
	\begin{equation}\label{eq:c_safety}
		c_{\mathrm{safety},ij}^{(k)}
		=
		w_{\mathrm{safety}}
		\left[
		\max
		\left(
		0,
		g_{\mathrm{safe}}
		-
		g_{\min,ij}^{(k)}
		\right)
		\right]^2 .
	\end{equation}
	Here, $w_{\mathrm{safety}}$ is the corresponding scalar weight. This cost penalizes candidates whose minimum predicted separation falls below $g_{\mathrm{safe}}$, and the penalty increases quadratically with the magnitude of the violation.
	
	\subsubsection{Overall Game-Aware Rollout Ranking}
	In addition to the game-aware cost, each sampled rollout is evaluated using a nominal quadratic tracking and control cost relative to a reference trajectory derived from the arc-length-parameterized raceline $\mathcal{G}$. For vehicle $i$, the reference progress $s_{i,t+\tau}^{r}$ is  propagated recursively along $\mathcal G$ using the associated reference velocity profile $v^r(s)$, starting from $s_{i,t}^{r}=s_{i,t}$:
	$$
	s_{i,t+\tau+1}^r = s_{i,t+\tau}^r + \Delta t \, v^r(s_{i,t+\tau}^r), \; \tau = 0, \dots, H-1.
	$$
	The corresponding reference states are obtained as  $\mathbf{x}_{i,t+\tau}^r = \mathbf{x}^r(s_{i,t+\tau}^r)$. For each candidate rollout $k$, the nominal tracking and control cost is defined as:
	\begin{equation}
		\begin{aligned}
			c_{\mathrm{track},i}^{(k)}
			=&
			\sum_{\tau=1}^{H}
			\left\|
			\mathbf{x}_{i,t+\tau}^{(k)}
			-
			\mathbf{x}^{r}_{i,t+\tau}
			\right\|_{\mathbf{Q}}^2
			\\
			&+
			\sum_{\tau=0}^{H-1}
			\left\|
			\mathbf{u}_{i,t+\tau}^{(k)}
			\right\|_{\mathbf{R}}^2
			+
			\sum_{\tau=1}^{H-1}
			\left\|
			\Delta \mathbf{u}_{i,t+\tau}^{(k)}
			\right\|_{\mathbf{S}}^2.
		\end{aligned}\label{eq:ctrack}
	\end{equation}
	Here $\Delta \mathbf{u}_{i,t+\tau}^{(k)}
	=
	\mathbf{u}_{i,t+\tau}^{(k)}
	-
	\mathbf{u}_{i,t+\tau-1}^{(k)}$, and $\mathbf{Q}$, $\mathbf{R}$, and $\mathbf{S}$ are weighting matrices. The total  cost assigned to candidate $k$ is:
	\begin{equation}\label{eq:allc}
		c_i^{(k)} = c_{\mathrm{track},i}^{(k)} + w_{\mathrm{game}} c_{\mathrm{game},i}^{(k)}.
	\end{equation}
	Here, the weight $w_{\mathrm{game}}$ balances the influence of the game-aware cost against  the nominal tracking and control cost.
	\vspace{-10pt}
	
	\subsection{Feasibility  Selection}\label{sec:selection}

	After batch cost evaluation, feasibility selection applies explicit track-boundary and inter-vehicle collision checks to the ranked candidates and selects the lowest-cost candidate that passes both checks.
	
	\subsubsection{Track-Boundary Check}
	
	The track-boundary clearance of candidate $k$ at prediction step $\tau$ is defined as follows:
	$$
	\rho_{\mathrm{bd},i,t+\tau}^{(k)}
	=
	\min
	\left(
	w^R\!(s_{i,t+\tau}^{(k)})
	+
	d_{i,t+\tau}^{(k)},
	\;
	w^L\!(s_{i,t+\tau}^{(k)})
	-
	d_{i,t+\tau}^{(k)}
	\right).
	$$
	The minimum boundary clearance over the prediction horizon is ${\rho}_{\mathrm{bd},i}^{(k)}
	=
	\min_{\tau=1,\ldots,H}
	\rho_{\mathrm{bd},i,t+\tau}^{(k)}.$  A candidate passes the track-boundary check  if its minimum clearance remains above the required threshold $ m_{\mathrm{bd}}$:
	\begin{equation}\label{eq:selection-boundary}
		{b}_i^{(k)} = \mathbb{I} \bigl[ {\rho}_{\mathrm{bd},i}^{(k)} \ge m_{\mathrm{bd}} \bigr].
	\end{equation}
	
	\subsubsection{Collision-Avoidance Check}
	
	The inter-vehicle collision check uses the time-aligned minimum-distance operator defined in~\eqref{eq:Dmin}.  
	For candidate rollout $k$, the minimum clearance over all opponents is defined as follows:
	$
	{\rho}_{\mathrm{obs},i}^{(k)}
	=
	\min_{j \in \mathcal{I}_{-i}}
	g_{\min,ij}^{(k)},
	$
	where $g_{\min,ij}^{(k)}$ is given by \eqref{eq:Dmin}.
	The candidate passes the collision-avoidance check if this minimum clearance
	remains above the collision threshold \(d_{\mathrm{col}}\):
	\begin{equation}\label{eq:selection-vehicle}
		{o}_i^{(k)}
		=
		\mathbb{I}
		\left[
		{\rho}_{\mathrm{obs},i}^{(k)}
		\ge
		d_{\mathrm{col}}
		\right].
	\end{equation}
	After ranking the candidate trajectories in $\mathcal{C}_{i,t}^{\ell}$ according to the cost in~\eqref{eq:allc}, each candidate is marked as feasible if it passes both the track-boundary and collision-avoidance checks.
	The lowest-cost candidate that passes both feasibility checks is selected as the planned trajectory at IBR iteration $\ell$:
	\begin{equation}\label{eq:selection}
		k_i^* = \arg\min_{k : h_i^{(k)}=1} c_i^{(k)}, h_i^{(k)} = b_i^{(k)} \, o_i^{(k)}.
	\end{equation}
	
	Feasibility selection complements cost-based rollout ranking: the composite cost ranks the candidates, whereas the boundary and collision checks determine which candidates satisfy the prescribed feasibility criteria.
	The selected control sequence is used to warm-start the next planning step.
	If no feasible candidate exists, a least-violation fallback strategy is invoked. The boundary and collision violations for each candidate are defined as:
	$
	\nu_{\mathrm{bd},i}^{(k)}
	=
	\max\bigl(0,m_{\mathrm{bd}}-\rho_{\mathrm{bd},i}^{(k)}\bigr), \, \nu_{\mathrm{obs},i}^{(k)} = \max\bigl(0, d_{\mathrm{col}} - \rho_{\mathrm{obs},i}^{(k)}\bigr),
	$
	with the total violation given by $\nu^{(k)}_i = \nu_{\mathrm{bd},i}^{(k)} + \nu_{\mathrm{obs},i}^{(k)}$. The fallback strategy selects the candidate with the smallest combined violation: $k^* = \arg\min_k \nu_i^{(k)}.$
	\vspace{-5pt}

	\section{Results and Discussion}\label{sec:results}
	
	This section evaluates SGTP in long-duration, closed-loop competitive multi-vehicle racing scenarios. We compare SGTP with a range of representative baselines, conduct ablation studies to assess the contributions of the proposed modules, and validate its performance across challenging tracks. The evaluation focuses on competitiveness and real-time efficiency.
	\vspace{-20pt}

	\subsection{Literature Benchmarks}\label{exp:baselines}

	The proposed method is compared with \textbf{eight literature benchmarks}. These include a sampling-based lattice planner~\cite{5980223}, the learning-based End2Race method in~\cite{qiao2025end2race}, and the conditional flow-matching (CFM) planner proposed in~\cite{NEURIPS2025_36d1e8aa}. End2Race~\cite{qiao2025end2race} predicts control commands from past velocities and LiDAR  inputs, whereas the CFM planner~\cite{NEURIPS2025_36d1e8aa} generates plans conditioned  on opponent predictions and the raceline.  We also compare SGTP with the Race Stack framework in~\cite{baumann2025forzaeth}, a rule-based FSM method validated in real-world racing, and with the EVO-MPCC method proposed in~\cite{li6127037evo}, which has also been validated in real-world racing scenarios. In addition, we include two MPPI-based baselines: standard MPPI~\cite{honda2025model} and Biased-MPPI~\cite{trevisan2024biased}. Finally, following~\cite{wang2021game}, we integrate EVO-MPCC into the IBR loop to compare SGTP with an MPC-based solver under the same IBR framework.

	Two ablation variants are evaluated. The first ablation, named SGTP w/o GC, removes the game-aware cost term in~\eqref{eq:game_cost} while retaining feasibility selection. The second, named GA-IBR-MPPI, incorporates the same game-aware cost into the IBR loop but uses the standard MPPI formulation, which updates the nominal control sequence through path-integral importance weighting~\cite{honda2025model}. In all cases, the planned trajectories are tracked with the same pure-pursuit controller~\cite{qiao2025end2race}. \revlzh{For simulation, we use the F1TENTH Gym environment described in~\cite{qiao2025end2race} and the vehicle model presented in~\cite{7995816}.}
	\vspace{-10pt}
	
	\subsection{Implementation and Racing Scenarios}\label{exp:setting}
	SGTP and the MPPI baselines use $K$=128 samples over a 1.2s prediction horizon ($H$=12), with $\sigma_a= 0.335$  and $\sigma_\delta=0.025$. The tracking-cost weights are $\mathbf{Q}$=$[60,60,47.75,39.48]$, $\mathbf{R}$=$[8.43,20.0]$, and $\mathbf{S}$=$[1.0,19.26]$. We set $w_{\mathrm{game}}$ =$60$,  $s_{\mathrm{contest}}$=8.0m,  $s_{\mathrm{role}}$=1.0m, $\mu_{\textrm{tail}}$=0.3,  and $g_{\mathrm{safe}}$=1.0m.  We use $L_{\mathrm{IBR}}$=$2$ IBR iterations, $m_{\mathrm{bd}}$ is 0.515m, and $d_{\mathrm{col}}$ is 0.9m. \revlzh{The game-aware cost weights $[w_{\mathrm{contest}}, w_{\mathrm{long}}, w_{\mathrm{block}}, w_{\mathrm{safety}}]$ = $[1.0,2.0,10.0,50.0]$ are manually tuned.} The MPC-based benchmark employs the CasADi~\cite{Andersson2019} software tool, with the IPOPT solver. We run our SGTP, the MPPI and learning-based baselines on an NVIDIA GeForce RTX 4060 GPU. 
	We employ the following $7$ racetracks from the open-source MapZoo\footnote{\href{https://github.com/zhouhengli/MapZoo}{https://github.com/zhouhengli/MapZoo}} repository: \textit{warehouse\_v1}, \textit{Berlin}, \textit{f-shape}, \textit{Brands Hatch}, \textit{Oschersleben}, \textit{MoscowRaceway}, and \textit{Nuerburgring}. For each track, the minimum-curvature raceline  \cite{heilmeier2020minimum} is used as the reference trajectory $\mathcal{G}$ for all vehicles. On each track, vehicles are initialized in a longitudinal queue along the track with a uniform 2m inter-vehicle spacing in ascending Frenet progress, with the ego vehicle always placed rearmost. The reference speeds of the opponent vehicles are scaled by a factor of 0.9 to promote close-range interactions.
	\vspace{-10pt}
	
	\subsection{Evaluation Metrics}\label{app:exp_Metrics}
	A trial is considered  a win if the ego vehicle achieves a greater Frenet progress gain (along the track reference line) than all opponents. 
	Specifically, the win indicator is defined as 
	$\mathbb{I}\left[\Delta s_e>\max_{j\neq e}\Delta s_j+\epsilon_s\right]$, 
	where $\Delta s_j=s_{j,T}-s_{j,0}$ is the progress gain of vehicle $j$, $\Delta s_e$ is the ego progress gain,
	$T$ is the trial duration, and $\epsilon_s=5\times10^{-3}$ is a small margin. 
	The win rate over all trials is reported as \textit{Wins}. 
	We also monitor the Collision-Free Win rate (\textit{CFW})  among winning trials; a higher value indicates cleaner wins without collisions. 
	Since the ego vehicle is initialized behind the opponents in each trial, we further report the Final Pass Ratio (\textit{FPR}) to quantify final overtaking performance.
	\textit{FPR} is defined as the fraction of opponents overtaken by the ego vehicle by the end of the trial:
	$
	\mathrm{FPR}
	=
	\frac{1}{n_v-1}
	\sum_{j\neq e}
	\mathbb{I}\left[s_{e,T}>s_{j,T}+\epsilon_s\right],
	$
	where $n_v$ is the number of vehicles. 
	To assess  long-duration racing performance, we report the maximum trial duration $D_{\max}$ and the mean trial duration $D_{\mathrm{mean}}$. A trial terminates when either a collision occurs or the predefined maximum duration is reached. The Close-interaction Segment Duration (\textit{CSD}) measures the mean duration of segments during which an opponent remains within $1.0\mathrm{m}$ longitudinally and $0.5\mathrm{m}$ laterally of the ego vehicle in Frenet coordinates. The mean ego velocity, denoted as $\bar{v}_{\text{ego}}$, is computed by averaging the ego speed over all samples from all trials. Mean Control Smoothness is reported by \textit{MCS} and is computed as $\frac{1}{T}\sum_{k=1}^{T}\|\Delta\mathbf{u}_k\|_2^2.$
	Computation Time (\textit{CT}), defined as the elapsed wall-clock time of each planning call, is used to evaluate real-time efficiency. For IBR-based game planners, \textit{CT} corresponds to the total  time over all IBR iterations and the additional ego response. We report $CT_{\mathrm{mean}}$, $CT_{\text{std}}$, and $CT_{\max}$ over all valid trials for statistical comparison.
	In the radar visualization of Fig.~\ref{fig:radar_multi_agents}, the six normalized axes represent long-duration interaction ($D_{\mathrm{mean}}$), winning ability (\textit{Wins}), overtaking ability (\textit{FPR}), close-range interaction (\textit{CSD}), aggressiveness ($\bar{v}_{\text{ego}}$), and runtime efficiency (the inverse-normalized $CT_{\mathrm{mean}}$).
	\vspace{-10pt}

	\begin{figure*}[!t]
		\centering
		{\includegraphics[width=7.0in]{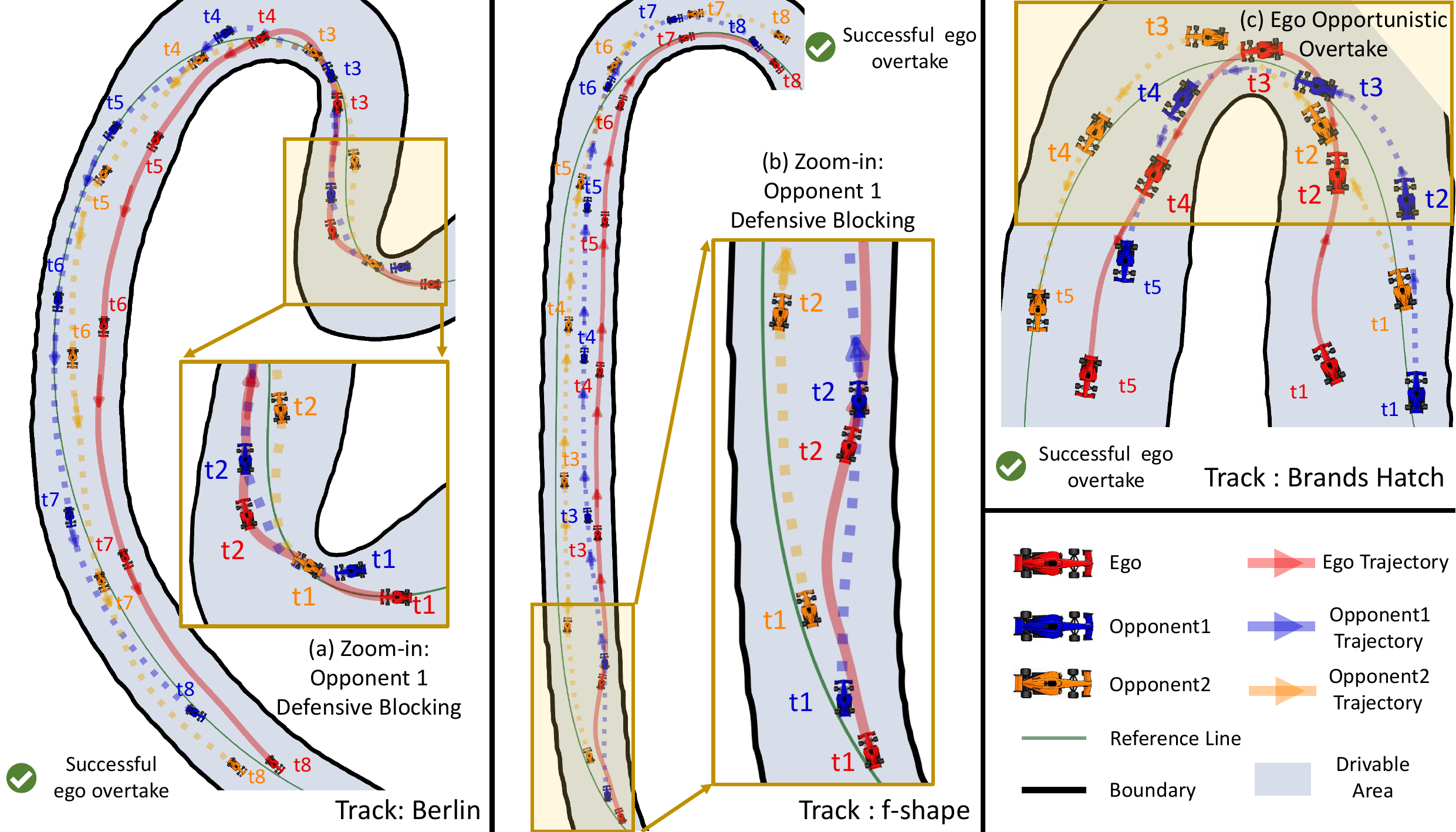}}
		\caption{\textbf{SGTP enables competitive behaviors and  reliable transitions between them in highly interactive scenarios on multiple racetracks.} The ego vehicle demonstrates diverse racing behaviors, including trailing and wheel-to-wheel contesting, before  completing an opportunistic overtake after  sustained close-range interactions. Videos are available on the project webpage.}
		\vspace{-15pt}
		\label{fig:vis_ot_block}
	\end{figure*}

	\subsection{Qualitative Evaluation of Multi-Behavior Interactions}
	
\begin{figure}[!t]
		\centering
		{\includegraphics[width=3.5in]{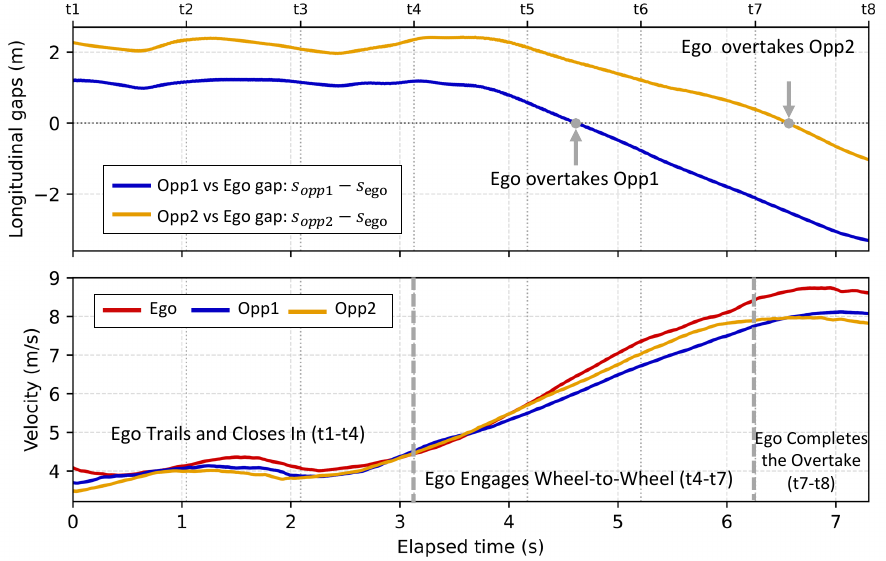}}
		\caption{
			Longitudinal gaps and velocity profiles from $t_1$ to $t_8$ on the \textit{Berlin} track (Fig.~\ref{fig:vis_ot_block}, left), where the SGTP-controlled ego vehicle overtakes two SGTP-controlled opponents.
		}
		\vspace{-10pt}
		\label{fig:progress_plot}
	\end{figure}

	Three challenging tracks, \textit{Berlin}, \textit{f-shape}, and \textit{Brands Hatch}, are used to evaluate SGTP in highly interactive racing scenarios. Each trial lasts up to 40s and includes three vehicles: the ego vehicle, Opponent 1 (O1), and Opponent 2 (O2). All vehicles are controlled by SGTP, and the opponents respond reactively and competitively to the ego vehicle during the race.
	As shown in Fig.~\ref{fig:vis_ot_block} and \ref{fig:progress_plot}, SGTP produces sustained close-range interactions in multi-vehicle racing. On the \textit{Berlin} track, O1 (\textcolor{blue}{blue}) executes a defensive blocking maneuver during time $t_1$--$t_2$ by deviating from the raceline (zoomed view (a)). This maneuver is to defend its position and prevent the ego vehicle (\textcolor{red}{red}) from overtaking. The ego vehicle responds by trailing O1 at $t_2$--$t_4$ and then uses its speed advantage (Fig.~\ref{fig:progress_plot}) to engage sustained wheel-to-wheel contesting  at $t_4$--$t_7$, progressively securing a favorable racing corridor and ultimately achieving an opportunistic overtake at $t_8$, as also shown in Fig.~\ref{fig:progress_plot}. On the \textit{f-shape} track (Fig.~\ref{fig:vis_ot_block}, middle), O1 initiates a defensive blocking maneuver at $t_2$ (see (b) zoom-in view). As the interaction evolves,  O1 moves farther from the raceline. The trade-off between blocking and raceline tracking then favors its return toward the raceline between $t_3$ and $t_5$. The ego preserves an overtaking corridor, maintains feasible wheel-to-wheel racing, and completes the overtake from $t_5$ to $t_8$. On the \textit{Brands Hatch} track, the ego vehicle simultaneously faces blocking from O2 ahead and overtaking pressure from O1 behind at $t_2$  (see (c) zoom-in view). Nevertheless, it completes an opportunistic overtake during $t_3$--$t_4$ by exploiting its speed advantage.

	These results show that SGTP generates diverse competitive behaviors through IBR updates and game-aware rollout ranking, including  blocking, wheel-to-wheel contesting, and opportunistic overtaking. The results further show that SGTP can transition smoothly among behaviors as interactions evolve.
	\vspace{-10pt}
	
	\subsection{Quantitative Results: Performance \& Real-Time Efficiency}
	For each of the 7 tracks in Section~\ref{exp:setting}, we sample 6 starting positions, yielding 42 trials with one ego and two opponents. Each trial runs for at most 50s and terminates if the ego vehicle is involved in a collision. \revlzh{The CFM and End2Race benchmarks use MPPI-controlled opponents for stable opponent predictions.} For all other methods, the ego vehicle and opponents use the same planner.

\begin{figure*}[htbp]
		\centering
		\includegraphics[width=6.8in]{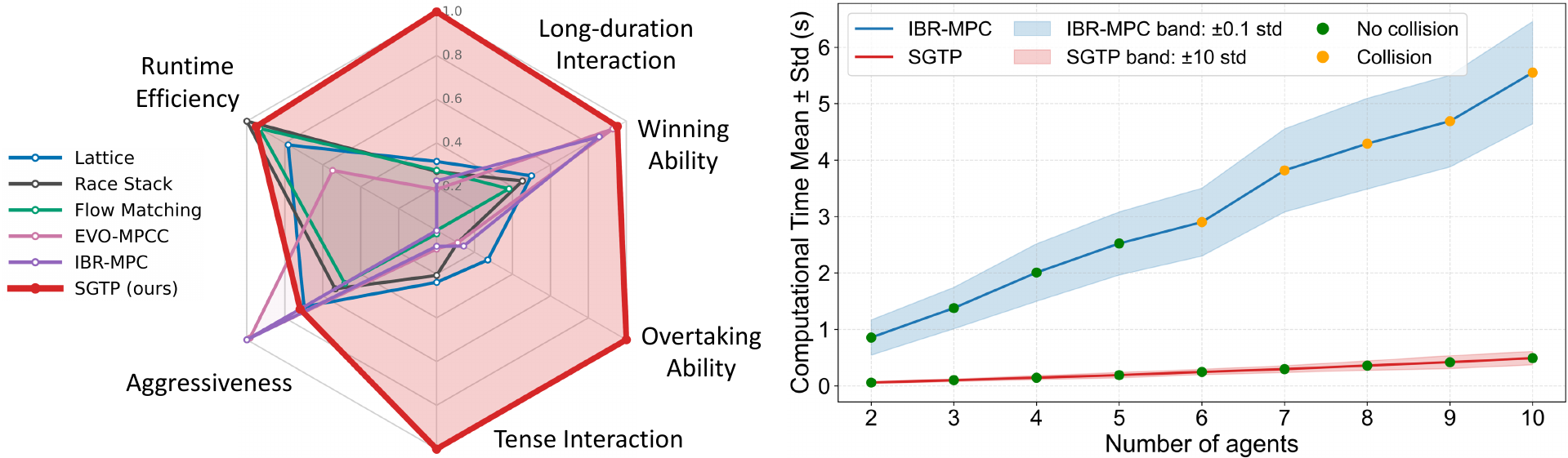}
		\caption{\textbf{Competitive performance \& scalability of SGTP.} Left plot: our SGTP achieves the best balance across the normalized metrics of Sec.~\ref{app:exp_Metrics}. Right plot: SGTP maintains low, consistent computation times and collision-free operation as the number of vehicles increases, outperforming IBR-MPC~\cite{wang2021game}. Our webpage shows videos with up to 10 agents.
		}
		\label{fig:radar_multi_agents}
		\vspace{-15pt}
	\end{figure*}

	Our SGTP achieves the best overall performance (Table~\ref{tab:batched_runs}), with \textbf{95.24\%} \textit{Wins} and \textbf{100.00\%} wins without collisions (\textit{CFW}). 
	It also achieves $D_{\mathrm{mean}} = 49.67$s under a predefined maximum trial duration of 50s, indicating that it can sustain racing for nearly the entire allowed duration. 
	In addition, our SGTP obtains the highest \textit{FPR} of \textbf{75.00\%} and \textit{CSD} of \textbf{0.99}, indicating successful overtaking after sustained close-range interactions.
	Meanwhile, it maintains competitive speed ($\bar{v}_{\text{ego}}$ = 5.86~m/s) and smooth controls (\textit{MCS} = 0.008). 
	As shown in Fig.~\ref{fig:radar_multi_agents}, SGTP presents the most balanced profile across long-duration racing, winning ability, overtaking ability, tense interaction, aggressiveness, and runtime efficiency. 
	It also runs in real time, with $CT_{\mathrm{mean}}=\mathbf{0.095}$~s (Table~\ref{tab:batched_runs}). Compared with the other benchmarks, only EVO-MPCC and IBR-MPC achieve relatively high win rates of 92.86\% and 85.71\%, respectively. However, neither method can sustain intense competitive interactions over extended racing periods. In addition, their planning processes are both time-consuming and unstable. Specifically, the $CT_{\mathrm{mean}}$ values for EVO-MPCC and IBR-MPC are 0.860s and 1.903s, while the corresponding $CT_{\text{std}}$ values are 0.646s and 1.043s, making these methods unsuitable for  real-time planning in rapidly changing multi-vehicle racing scenarios.
	The ablation results verify the importance of our two key modules: game-aware cost and feasibility selection. 
	Without the game-aware cost, SGTP w/o GC drops to 50.00\% \textit{Wins} and 14.29\% \textit{CFW}, indicating weaker competitive  performance. 
	Without the  feasibility selection, GA-IBR-MPPI achieves only 57.14\% \textit{Wins} and 0.00\% \textit{CFW}, while its  $D_{\max}$ is only 5.21s, showing that  feasibility selection is critical for long-duration competitive racing.

	\begin{table*}[htbp]
		\centering
		\caption{Quantitative Comparison and Ablation Study of Multi-Vehicle Racing Planners, Demonstrating the Effectiveness of the Proposed SGTP in Long-duration Competitive Racing and  Real-Time Planning.}
		\begin{adjustbox}{max width=18cm}
			\large
			\renewcommand{\arraystretch}{1.05}
			
			\begin{tabular}{r|c|c|c|ccccccc|ccc}
				\toprule
				\multicolumn{2}{c|}{\multirow{2}[4]{*}{Methods}} & \multirow{2}[4]{*}{Wins$^a \uparrow$} & \multirow{2}[4]{*}{CFW$\uparrow$} & \multicolumn{7}{c|}{Long-duration Competitive Performance $\uparrow$} & \multicolumn{3}{c}{Real-Time Efficiency $\downarrow$} \\
				\cmidrule{5-14}    \multicolumn{2}{c|}{} &       &   & FPR$^a$    & CSD$^b$   & $D_{\mathrm{mean}}^c$ &  $D_{\mathrm{ratio}}^c$ & $D_{\max}^c$  & $\bar{v}_{\text{ego}}$   & MCS$^c \downarrow$   &  \large{$CT_{\mathrm{mean}}$} & \large{$CT_{\text{std}}$} &  \large{$CT_{\max}$} \\
				\midrule
				\midrule
				\multicolumn{1}{c|}{Sampling} & Lattice\cite{5980223} & 50.00\% & 0.00\% & 20.24\% & \underline{0.23}  & 16.73  & 33.46\% & \underline{44.02} & 5.75  & 0.928  & 0.415  & 0.072  & 0.560  \\
				\midrule
				\multicolumn{1}{c|}{FSM} & Race Stack\cite{baumann2025forzaeth} & 45.24\% & 5.26\% & 8.33\% & 0.20  & 14.51  & 29.02\% & \textbf{50}    & 4.90  & 0.037  & -  & -  & -  \\
				\midrule
				\multicolumn{1}{c|}{\multirow{2}[2]{*}{Learning}} & End2Race\cite{qiao2025end2race} & 66.67\% & 0.00\% & 11.90\% & 0.05  & 2.84  & 5.69\% & 4.48  & 4.02  & \textbf{0.002}  & \textbf{0.003}  & \textbf{3.9e-4} & \textbf{0.004}  \\
				& {CFM}\cite{NEURIPS2025_36d1e8aa} & 38.10\% & 0.00\% & 0.00\% & 0.02  & 14.78  & 29.55\% & \textbf{50}    & 4.63  & 0.120  & 0.127  & 0.012  & 0.141  \\
				\midrule
				\multicolumn{1}{c|}{\multirow{3}{*}{\makecell[c]{Predictive\\Control}}} & MPPI\cite{honda2025model}  & 4.76\% & 0.00\% & 3.57\% & 0.00  & 1.60  & 3.20\% & 5.15  & 2.16  & \underline{0.004}  & \underline{0.041}  & \underline{0.001}  & \underline{0.042}  \\
				& Biased-MPPI\cite{trevisan2024biased} & 2.38\% & 0.00\% & 2.38\% & 0.00  & 1.36  & 2.73\% & 2.7   & 1.90  & \textbf{0.002}  & 0.095  & 0.005  & 0.099  \\
				& EVO-MPCC\cite{li6127037evo} & \underline{92.86\%} & 2.56\% & 8.33\% & 0.08  & 10.58  & 21.17\% & \textbf{50}    & \underline{7.23}  & 0.011  & 0.860  & 0.646  & 3.743  \\
				\midrule
				& IBR-MPC\cite{wang2021game} & 85.71\% & 5.56\% & 10.71\% & 0.07  & 12.46  & 24.91\% & \textbf{50}    & \textbf{7.32}  & 0.012  & 1.903  & 1.043  & 4.153  \\
				\multicolumn{1}{c|}{\multirow{2}[2]{*}{Game}} & SGTP w/o GC & 50.00\% & \underline{14.29\%} & \underline{28.57\%} & 0.14  & \underline{20.10}  & \underline{40.20\%} & \textbf{50}    & 4.87  & 0.006  & 0.116  & 0.006  & 0.123  \\
				& GA-IBR-MPPI & 57.14\% & 0.00\% & 19.05\% & 0.02  & 2.22  & 4.44\% & 5.21  & 2.86  & 0.007  & 0.078  & 0.003  & 0.084  \\
				& \textbf{SGTP (ours)} & \cellcolor[rgb]{ .949,  .949,  .949}\textbf{95.24\%} & \cellcolor[rgb]{ .949,  .949,  .949}\textbf{100.00\%} & \cellcolor[rgb]{ .949,  .949,  .949}\textbf{75.00\%} & \cellcolor[rgb]{ .949,  .949,  .949}\textbf{0.99 } & \cellcolor[rgb]{ .949,  .949,  .949}\textbf{49.67}  & \cellcolor[rgb]{ .949,  .949,  .949}\textbf{99.35\%} & \cellcolor[rgb]{ .949,  .949,  .949}\textbf{50} & \cellcolor[rgb]{ .949,  .949,  .949}5.86  & \cellcolor[rgb]{ .949,  .949,  .949}0.008  & \cellcolor[rgb]{ .949,  .949,  .949}0.095 & \cellcolor[rgb]{ .949,  .949,  .949}0.004  & \cellcolor[rgb]{ .949,  .949,  .949}0.102  \\
				\bottomrule
			\end{tabular}%
			
		\end{adjustbox}
		
		\begin{tablenotes}
			\footnotesize
			\item \textbf{Bold} and \underline{underlined} values indicate the best and second-best results, respectively; shading highlights SGTP. $\uparrow$/$\downarrow$ denote higher/lower is better.
			
			\item $^a$ \textit{Wins} denotes the progress-based win rate, and \textit{CFW} denotes the collision-free win rate. \textit{FPR} denotes the Final Pass Ratio, i.e., the fraction of opponents overtaken by the ego vehicle by the end of a trial based on cumulative Frenet progress. Higher values are better for all three metrics.
			
			\item $^b$ \textit{CSD} measures the mean duration of close-interaction segments. In conjunction with $D_{\mathrm{mean}}$, a larger \textit{CSD} indicates that high-intensity close-range interactions are sustained over longer periods.
			
			\item $^c$  $D_{\mathrm{mean}}$ and $D_{\max}$ quantify the durations of sustained racing interactions, with larger values indicating longer  interactions. $D_{\mathrm{ratio}}$ is computed as $D_{\mathrm{mean}}/\text{(maximum trial time)}$ to represent the race completion ratio. MCS denotes the mean control smoothness.

		\end{tablenotes}
		\label{tab:batched_runs}%
		\vspace{-10pt}
	\end{table*}%

	\vspace{-10pt}
	
	\subsection{Defensive Capability and Robustness to Perturbations}
	
	Across the 7 tracks in Section~\ref{exp:setting}, we sample 15 positions per track, yielding 105 trials to evaluate SGTP’s defensive capability and robustness to prediction perturbations (Table~\ref{tab:small_table}). \revlzh{To evaluate defensive capability, the rearmost vehicle uses EVO-MPCC~\cite{li6127037evo}, while both opponents use SGTP, and their performance is reduced by 10\% to complicate the defense. Only the EVO-MPCC agent performs the additional final response.} Despite this advantage, its CFW is only 3.88\%, demonstrating that SGTP can effectively defend against an MPC-based planner without performing the final response.

	We next evaluate SGTP using perturbed opponent-trajectory predictions generated by the perturbation model in~\cite{zhang2022adversarial}. In this evaluation, the two opponents and the ego vehicle (starting from behind) are controlled by SGTP.  Under mild perturbations, $D_{\mathrm{mean}}$ decreases by only 0.52\%, while \textit{CSD} decreases by 55.88\%, indicating more conservative interactions. This results from the safety cost and  feasibility selection, which penalize or reject candidates with insufficient clearance. Under severe perturbations, \textit{CSD} and $D_{\mathrm{mean}}$ decrease by 63.86\% and 4.41\%, respectively. Nevertheless, SGTP maintains a CFW of 94.12\%, demonstrating its capability to avoid collisions even with noisy opponents' predictions.
	
	\vspace{-10pt}

	\subsection{Scalability Analysis Across Vehicle Counts}

	\begin{table}[!t]
		\centering
		\caption{SGTP’s capability to defend \& block (first row) and robustness to opponents' prediction uncertainty (last 3 rows).}
		\begin{adjustbox}{max width=8.5cm}
			\begin{tabular}{c|ccccr}
				\toprule
				Methods & $D_{\mathrm{mean}}$ & $\Delta D_{\mathrm{mean}}$  & CSD   & $\Delta \mathrm{CSD}$  & \multicolumn{1}{l}{CFW} \\
				\midrule
				\midrule
				EVO-MPCC vs SGTP & 12.00  & -     & 0.06  & -     & 3.88\% \\
				SGTP w/o noise & 49.08  & -     & 1.01  & -     & 96.12\% \\
				SGTP mild noise & 48.82  & -0.52\% & 0.44  & -55.88\% & 97.14\% \\
				SGTP severe noise & 46.91  & -4.41\% & 0.36  & -63.86\% & 94.12\% \\
				\bottomrule
			\end{tabular}%
		\end{adjustbox}
		\label{tab:small_table}%
		\begin{tablenotes}
			\footnotesize
			\item $\Delta D_{\mathrm{mean}},(\%)$ and $\Delta \mathrm{CSD},(\%)$ are computed relative to the corresponding values for SGTP w/o noise. The mild and severe settings use $(\sigma_{\mathrm{lon}},\sigma_{\mathrm{lat}})=(0.03,0.015)$ and $(0.08,0.04)$, respectively.
		\end{tablenotes}
	\end{table}%

	We now assess the scalability of SGTP as the number of interacting vehicles increases. On the \textit{Brands Hatch} track, we vary the number of vehicles from 2 to 10, and each trial lasts 11s. 
	As shown in Fig.~\ref{fig:radar_multi_agents}, SGTP maintains low and stable computational time with small variance, while remaining collision-free across all vehicle counts. 
	In contrast, IBR-MPC exhibits rapidly increasing computational time and greater variance, with collisions occurring in denser scenarios. 
	Overall, SGTP combines IBR reasoning with sampling-based planning to maintain real-time performance and generate reliable  behaviors across varying numbers of vehicles.
	\vspace{-5pt}

	\section{Conclusions}\label{sec:conclusion}

	This letter presented SGTP, a real-time game-theoretic planner for competitive multi-vehicle racing. By exploiting GPU-parallel  sampling and rollout evaluation, SGTP efficiently approximates unilateral best responses in IBR. Feasibility selection explicitly enforces the track-boundary and collision-avoidance constraints, supporting reliable transitions between diverse behaviors. The key finding is that the game-aware cost  can produce diverse competitive behaviors, including blocking, wheel-to-wheel contesting, and overtaking. Future work will reduce parameter tuning and incorporate world-model-based opponent style recognition to improve prediction.
	
	\vspace{-5pt}

	\bibliographystyle{IEEEtran}
	\bibliography{bibref}
	
	\clearpage
	
\end{document}